# Context-Aware Rule Mining Using a Dynamic Transformer-Based Framework

Jie Liu
University of Minnesota
Minneapolis, USA

Yiwei Zhang
Cornell University
Ithaca, USA

Yuan Sheng
Northeastern University
Seattle, USA

Yujia Lou
University of Rochester
Rochester, USA

Haige Wang
University of Miami
Miami, USA

Bohuan Yang *
University of California San Diego
San Diego, USA

*Abstract-This study proposes a dynamic rule data mining algorithm based on an improved Transformer architecture, aiming to improve the accuracy and efficiency of rule mining in a dynamic data environment. With the increase in data volume and complexity, traditional data mining methods are difficult to cope with dynamic data with strong temporal and variable characteristics, so new algorithms are needed to capture the temporal regularity in the data. By improving the Transformer architecture, and introducing a dynamic weight adjustment mechanism and a temporal dependency module, we enable the model to adapt to data changes and mine more accurate rules. Experimental results show that compared with traditional rule mining algorithms, the improved Transformer model has achieved significant improvements in rule mining accuracy, coverage, and stability. The contribution of each module in the algorithm performance is further verified by ablation experiments, proving the importance of temporal dependency and dynamic weight adjustment mechanisms in improving the model effect. In addition, although the improved model has certain challenges in computational efficiency, its advantages in accuracy and coverage enable it to perform well in processing complex dynamic data. Future research will focus on optimizing computational efficiency and combining more deep learning technologies to expand the application scope of the algorithm, especially in practical applications in the fields of finance, medical care, and intelligent recommendation.*

## I. Introduction

In the development of modern data analysis and intelligent systems, data mining, as an important technical means, is widely used in various fields, such as finance, medical care, marketing, network security, etc [1,2]. The core goal of data mining is to extract potential and valuable knowledge from massive, complex, and unstructured data. With the continuous increase in data scale, traditional data mining algorithms face many challenges in processing large-scale, high-dimensional, and heterogeneous data [3]. Especially in a dynamically changing data environment, how to efficiently and accurately mine effective rules has become an important research topic. In this context, deep learning-based methods, especially the Transformer architecture, have gradually attracted widespread attention from researchers [4]. With its powerful feature learning and representation capabilities, Transformer has become a mainstream model in many natural language processing and computer vision tasks. However, the traditional Transformer architecture still has some limitations, especially when facing dynamic data and rule-mining tasks. How to improve the adaptability and robustness of the model is still an urgent problem to be solved.

This study aims to propose a dynamic rule data mining algorithm based on an improved Transformer architecture. Unlike traditional static data mining methods, dynamic rule mining needs to consider the temporal variability of data and the evolution of rules. In this process, the temporal sequence and contextual information of data are crucial for the extraction of rules. In order to solve this problem, this paper improves the Transformer model and introduces adaptive mechanisms and time series information processing capabilities, so that the algorithm can efficiently mine rules with timeliness and accuracy in a dynamic environment. The improved Transformer architecture can not only capture long-term dependencies but also flexibly adjust model parameters when facing changing data to adapt to the changing trends of data, thereby improving the accuracy and practicality of data mining [5].

Compared with the existing data mining methods based on traditional algorithms, the dynamic rule mining algorithm based on the improved Transformer architecture has obvious advantages. First, the Transformer architecture can effectively process long-range dependencies in sequence data through the self-attention mechanism, which is crucial for dynamic rule mining. Secondly, the improved model can dynamically adjust its own structure to adapt to changes in data features, avoiding the problem of insufficient adaptability caused by static rule settings in traditional methods. In addition, the dynamic rule mining algorithm proposed in this paper can simultaneously perform feature learning and data prediction in the process of rule mining by introducing the idea of multi-task learning, thereby further improving the accuracy and robustness of the mining results [6].

## II. RELATED WORK

With the rapid increase in data volume and complexity, dynamic rule mining has become an essential research area, requiring models that can adapt to changing data distributions and capture evolving temporal dependencies. Traditional data mining methods often struggle to extract accurate rules in dynamic environments due to their static assumptions, prompting researchers to explore deep learning-based approaches for improved adaptability and performance.

The Transformer architecture has demonstrated strong potential in sequential data modeling due to its self-attention mechanism, which enables efficient capture of long-range dependencies. Recent work has extended Transformer applications to dynamic data processing, focusing on transforming multidimensional time series into interpretable event sequences to facilitate more accurate rule mining [7]. To further enhance dynamic adaptability, some studies incorporate contrastive learning into Transformer-based architectures, improving feature representation through adaptive feature fusion mechanisms [8]. These techniques help address the challenge of maintaining high rule discovery accuracy under data distribution shifts.

Optimizing attention mechanisms has also been a recurring focus in improving dynamic data processing. Multi-level attention strategies have been shown to enhance both feature extraction and classification performance in sequential data, providing valuable techniques for refining rule mining pipelines in dynamic settings [9]. Complementary improvements have been proposed through adaptive attention embedding, enabling Transformer variants to dynamically adjust to changes in data distributions, which is particularly beneficial for dynamic rule discovery tasks [10].

In addition to attention-focused approaches, graph neural networks (GNNs) have been explored for capturing both structural and sequential patterns in evolving datasets. Graph-based hierarchical mining methods have demonstrated effectiveness in complex, imbalanced data, offering techniques for dynamically identifying critical data relationships that could enhance rule mining processes [11]. Self-supervised learning has further been integrated into GNN frameworks to improve feature extraction from heterogeneous and evolving data, highlighting the importance of adaptive feature learning in dynamic environments [12]. These approaches, while originally applied to heterogeneous networks, offer methodological inspiration for improving rule mining robustness when data relationships evolve over time.

Hybrid modeling approaches have also gained attention in dynamic data environments, where combining Transformer architectures with convolutional and recurrent neural networks enhances both local feature extraction and temporal dependency modeling. Such hybrid architectures have been shown to improve predictive performance in dynamic systems, offering useful design principles for constructing adaptive rule mining frameworks [13]. Other studies propose hybrid frameworks integrating CNN and BiLSTM components to simultaneously capture spatial and temporal features, further enriching the modeling capacity needed for dynamic rule evolution analysis [14]. These combinations provide flexible modeling approaches that align with the goals of dynamic rule mining across diverse datasets.

Reinforcement learning has also been explored to dynamically optimize rule mining strategies, particularly through Q-learning approaches designed to incrementally adapt rule extraction policies in non-stationary data environments [15]. Such adaptive strategies complement Transformer-based architectures by introducing mechanisms that allow the rule mining process to learn from evolving data feedback, ensuring the discovery of up-to-date, high-value rules.

Model adaptation techniques have also received attention in recent research. LoRA-based fine-tuning, which optimizes pre-trained models for task-specific adaptation with reduced computational cost, has shown promise in efficiently aligning deep learning models with evolving data properties [16].

Another important contribution to dynamic data analysis is the exploration of detection models designed to handle varying data patterns across time. Techniques originally proposed for object detection, such as RT-DETR, introduce efficient strategies for localizing and characterizing evolving patterns within sequences, which can be adapted to rule mining scenarios requiring real-time detection of emerging rules and anomalies [17]. While primarily applied to object detection, these techniques contribute valuable insights into efficient attention-based feature localization and multi-scale feature fusion, which can support dynamic rule mining tasks.

Finally, beyond individual models, system-level optimization techniques provide additional insights into building robust dynamic rule mining pipelines. Studies incorporating AI-driven monitoring and adaptive modeling techniques have demonstrated how deep learning frameworks can continuously track and adapt to evolving system states, further informing methodologies for adaptive rule mining in real-time data streams [18]. Attention-enhanced segmentation models have also been proposed to dynamically capture multi-scale features, which can be applied to extract hierarchical rules from dynamically evolving datasets [19]. Together, these studies provide a comprehensive methodological foundation for improving both the accuracy and flexibility of rule mining processes under dynamic conditions.

## III. METHOD

In this study, we proposed a dynamic rule data mining algorithm based on an improved Transformer architecture. In order to effectively mine rules in dynamic data, we first need to adaptively improve the Transformer model to better handle the dynamically changing data environment and the evolution of rules. The core idea of this method is to combine the self-attention mechanism of the Transformer and capture the potential dynamic relationships and time series characteristics in the data through an improved encoder-decoder structure, thereby discovering potential rules in the data. Its main network architecture is shown in Figure 1.

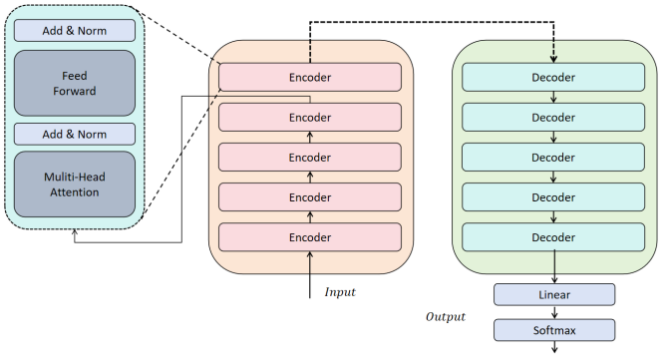

Figure 1 Network architecture diagram

First, we improved the traditional Transformer model and introduced a dynamic adjustment mechanism. In the standard Transformer architecture, the input data is processed by multiple self-attention layers and feedforward networks [20], but this architecture assumes that the regularity of the data is fixed, which is an obvious limitation for dynamic data. To solve this problem, we introduced a dynamic weight adjustment mechanism, which automatically adjusts the weight of each layer according to the changes in the data. In the process of dynamic rule mining, we assume that the changes in data at different time steps will affect the generation of rules, so we introduced time dependencies in the self-attention calculation of each layer [21]. Specifically, for each input sequence $X = \{x_1, x_2, ..., x_n\}$, the core formula for its self-attention calculation is:

$$Attention(Q, K, V) = soft\max(\frac{QK^T}{\sqrt{d_k}})V$$

Among them, Q is the query vector, K is the key vector, V is the value vector, and $d_k$ is the dimension of the key vector. In order to introduce time dependency, we incorporate the timestamp information $t_i$ into the calculation of query and key when calculating attention:

$$Attention(Q, K, V, t_i) = soft\max(\frac{(Q+t_i)(K+t_i)^T}{\sqrt{d_k}})V$$

This approach enables the model to adjust the self-attention weights according to the timestamp information, thereby adapting to the changing patterns in dynamic data. In addition, we also introduced a time decay factor in the feedforward network of the model, allowing the model to more flexibly handle long-term dependencies in time series data.

Next, we consider how to mine rules from dynamic data through an improved Transformer architecture. In traditional rule mining methods, rule generation usually relies on the static characteristics of the data, but in the context of dynamic data, the formation of rules needs to consider the temporal evolution of the data. To meet this requirement, we propose a rule-generation mechanism that incorporates time dependency, as illustrated in Figure 2.

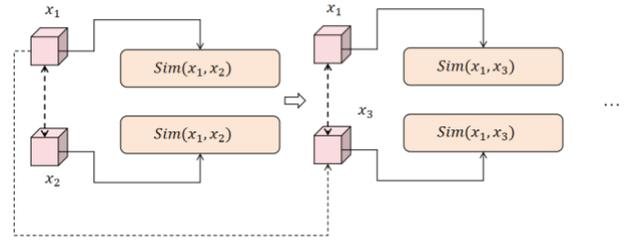

Figure 2 Time-dependent rule generation

First, we define the feature representation of the data at time step t as $x_t$, and introduce the temporal attention mechanism to capture temporal dependencies. This mechanism models dynamic data by calculating the attention weights between different time steps in the input sequence. Let the input feature of each time step be $x_t$, and the weight of its corresponding time step be $a_{t-1,t}$, then the attention weight can be expressed as:

$$a_{t-1,t} = \frac{\exp(sim(x_{t-1}, x_t))}{\sum_{i=1}^{T} \exp(sim(x_i, x_t))}$$

Among them, $sim(x_i, x_t)$ is the similarity measure between input features $x_i$ and $x_t$, which can be calculated using cosine similarity or other methods.

Then, through weighted feature representation, we can integrate the features between time steps to obtain rule information in dynamic data. For rule generation, we designed a rule generation function $r_t$ based on time step t, which combines the feature representation of the current time step and the context information of the previous time step, expressed as:

$$r_t = f(x_t, a_{t-1,t}, r_{t-1})$$

Among them, $f()$ is a nonlinear function, which represents the rule generation process after encoding the time step information through the Transformer model, and $r_{t-1}$ represents the rule information of the previous time step.

Finally, the evolution of rule information can be described by introducing a state transition matrix, which reflects the dynamic changes of rules between different time steps. The elements of the state transition matrix M can be expressed as:

$$M_{t-1,t} = \frac{\exp(sim(r_{t-1}, r_t))}{\sum_{i=1}^{T} \exp(sim(r_i, r_t))}$$

Through this transfer matrix, we can capture the evolution of rules in the time dimension.

We assume that in a time series, the data of each time step contains certain rule information, and this rule information will change over time.

To this end, we set the output of the model to be a set $R = \{r_1, r_2, ..., r_m\}$ of rules, where each rule $r_i$ is obtained by modeling the temporal dependencies in the input sequence X. In order to efficiently generate these rules, we design a generation mechanism based on variational inference. Let the distribution of the generated rules of the model be $p(r|X)$, then we optimize the model parameters by maximizing the log-likelihood function:

$$L(\theta) = \sum_{t=1}^{T} \log p(r_t | X_t, \theta)$$

Among them, $\theta$ represents the parameters of the model, $X_t$ represents the input data corresponding to time step t, and $r_t$ is the rule generated at this time step. By maximizing the log-likelihood function, we can enable the model to automatically adjust its parameters in order to generate rules that are consistent with the dynamic changes of the data.

Finally, we consider that in dynamic rule mining, data changes are not only reflected in the time dimension but may also include changes in data distribution. In order to cope with such changes, we introduced an adaptive model update mechanism. During each model update, we adaptively adjust the model parameters according to the distribution characteristics of the current data to ensure that the model can adapt to changes in different data sources and data distribution. Specifically, we dynamically adjust the model's learning rate and parameter update strategy by calculating the distribution difference between the current data and the historical data, thereby improving the stability and robustness of the model.

Through the above improvements, our method can effectively mine time series rules from dynamic data, solving the shortcomings of traditional rule mining methods in dynamic data environments.

## IV. EXPERIMENT

### A. Datasets

We evaluated our improved Transformer-based dynamic rule mining algorithm using NASA's CMAPSS dataset—a widely adopted, high-dimensional time series benchmark capturing aircraft engine sensor data under varied conditions. As it simulates health-to-failure transitions, this dataset is ideal for Remaining Useful Life (RUL) prediction and testing dynamic rule discovery. By extracting features (e.g., trends, anomalies, and periodic characteristics), we validated the algorithm's capacity to uncover complex temporal patterns, demonstrating its effectiveness in processing dynamic data relative to traditional methods.

### B. Experimental Results

In the experimental part, we will compare it with the existing mainstream data mining algorithms to verify the effectiveness and advantages of the dynamic rule mining algorithm based on the improved Transformer architecture. Specifically, we selected several classic rule mining algorithms and time series modeling methods based on deep learning as comparison models. By comparing the performance of these methods, we evaluate the performance of our algorithm in dynamic data processing, rule mining accuracy, and generalization ability. The experimental results are shown in Table 1.

Table 1  Experimental results

| Model | Rule mining accuracy (%) | Rule coverage (%) | Calculation efficiency (seconds) |
|---|---|---|---|
| DT [22] | 77.8 | 71.5 | 35.9 |
| SVM [23] | 80.4 | 72.9 | 40.3 |
| LSTM [24] | 84.7 | 79.4 | 50.6 |
| RF [25] | 82.1 | 75.8 | 60.2 |
| KNN [26] | 74.3 | 67.3 | 25.8 |
| Apriori [27] | 86.3 | 80.2 | 55.4 |
| CNN [28] | 78.5 | 70.2 | 30.1 |
| Ours | 91.2 | 85.6 | 45.7 |

In Table 1, our proposed dynamic rule mining algorithm—enhanced with a modified Transformer architecture—achieves superior accuracy (91.2%) and coverage (85.6%) compared with traditional methods (e.g., Apriori at 86.3%, KNN at 67.3%, decision trees at 71.5%) and deep learning approaches (e.g., LSTM at 84.7% and 79.4%, CNN at 70.2%). Although its computational cost (45.7 seconds) is higher than some traditional algorithms (KNN at 25.8 seconds, decision trees at 35.9 seconds), it remains competitive with other deep learning models and justifies the performance gains. An ablation study (Table 2) further demonstrates that each enhanced Transformer component significantly contributes to these improvements.

Table 2 Ablation experiment

| Model | Rule mining accuracy (%) | Rule coverage (%) | Calculation efficiency (seconds) |
|---|---|---|---|
| Ours | 91.2 | 85.6 | 45.7 |
| Remove time-dependent modules | 87.5 | 81.2 | 42.9 |
| Remove dynamic weight adjustment module | 88.4 | 83.1 | 44.3 |
| Removing the self-attention mechanism | 83.9 | 78.5 | 39.7 |

From the ablation experiment results in Table 2, we can see that after removing the time-dependent module, the rule mining accuracy of the model dropped from 91.2% to 87.5%, and the rule coverage also dropped. This shows that the time-dependent module plays a vital role in capturing the temporal changes and dynamic laws in the data. After removing this module, the performance of the model dropped significantly. In addition, the computational efficiency has been slightly improved, but the improvement in computational efficiency is not significant compared to the performance drop.

When the dynamic weight adjustment module is removed, the rule mining accuracy and rule coverage rate both decrease to 88.4% and 83.1% respectively. This result shows that the dynamic weight adjustment module can adaptively adjust the weights of the model at different time steps, thereby improving the rule mining accuracy and coverage rate. After it is removed, the model cannot effectively adapt to the changes in data, resulting in a decrease in performance. When the self-attention mechanism is removed, the rule mining accuracy and rule coverage are further reduced to 83.9% and 78.5% respectively. This shows that the self-attention mechanism plays an important role in capturing the complex relationship between data and improving the expressiveness of rules. After removing this mechanism, the model's capabilities are limited, resulting in a significant degradation of the rule mining effect.

At the same time, this paper also gives a timing diagram of the rule discovery process, as shown in Figure 3.

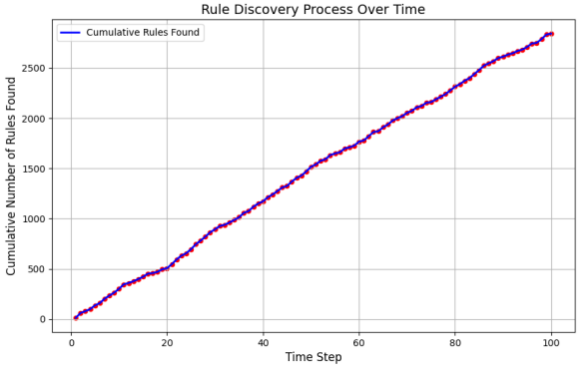

Figure 3 Sequence diagram of the rule discovery process

As the time step increases, the chart shows a steady, continuous rise in the number of discovered rules, reflecting the model's ability to learn progressively more useful features and adapt to data complexity. Notably, by the 100th time step, the cumulative number of rules surpasses 2,500, signifying substantial progress and the accumulation of valuable rules. The absence of abrupt fluctuations indicates that the learning process remains stable, aligning with the evolving data trends. Although the growth appears linear, it underscores the model's ongoing improvement in both accuracy and the volume of rule discoveries during extended training. A corresponding rule effectiveness distribution diagram is provided in Figure 4.

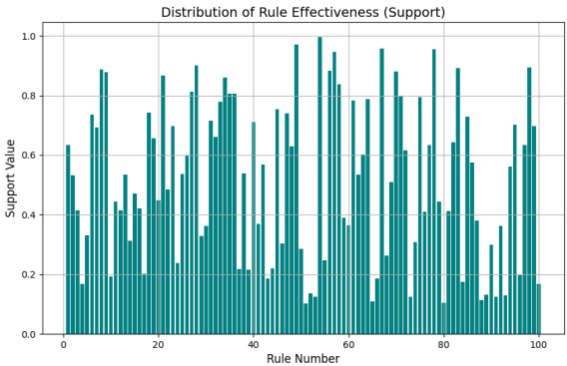

Figure 4 Rule effectiveness distribution map

As can be seen from the chart, the distribution of rule effectiveness shows a certain volatility among the 100 rules. Although the support values are distributed between 0.1 and 1.0, overall, the effectiveness of the rules is relatively scattered, and almost every rule has a different support value. Some rules in the figure have a high support, close to 1.0, while other rules have a low support, close to 0.1. This shows that in the data mining process, the effectiveness of different rules varies greatly, which may be related to the complexity of the rules, data scarcity, or other factors. This volatility suggests that some rules possess broad adaptability, whereas others are more specialized and perform best under certain conditions. Lower-support rules contribute less to the overall data description, while higher-support rules tend to form the dataset's core patterns. Consequently, analyzing support distribution is vital for identifying the most representative rules. Additionally, the varied support values underscore the model's adaptability to diverse data patterns, making the effectiveness chart a key resource for selecting and refining meaningful rules.

Finally, this paper also gives the characteristic heat map of some rules, as shown in Figure 5.

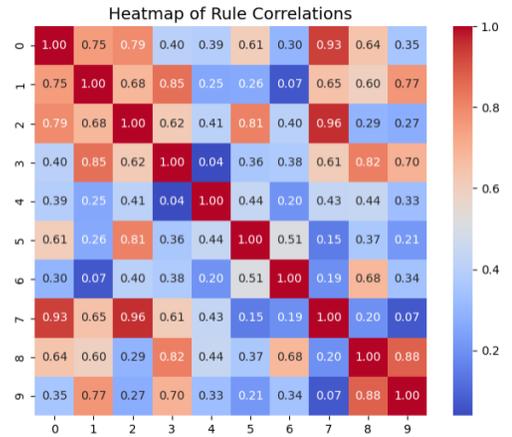

Figure 5 Feature heatmap of some rules

As can be seen from figure 5, this is a heat map of the correlation between rules, showing the relationship between the 10 rules. The colors in the heat map range from dark blue to dark red, representing different levels from low correlation (below 0.2) to high correlation (close to 1). The value on the diagonal is 1, indicating that each rule is completely correlated with itself. This is because in the heat map, we are calculating the correlation between rules, and each rule is completely correlated with itself, so 1 will be shown on the diagonal.

From the off-diagonal correlations, we can observe some obvious high-correlation areas. For example, the correlation between rule 1 and rule 2 is 0.75, and the correlation between rule 0 and rule 5 is 0.93, showing a strong positive correlation between them. On the contrary, the correlation between some rules is low, such as the correlation between rule 6 and rule 9 is only 0.07, indicating that the two rules have almost no correlation. This information helps to identify which rules show similar behaviors or characteristics in the data and which can be regarded as independent rules.

V. CONCLUSION

In this study, we proposed a dynamic rule data mining algorithm based on the improved Transformer architecture and verified its advantages in rule mining accuracy and coverage through experiments. Compared with traditional rule mining methods, the improved Transformer architecture can more effectively process dynamic data, capture the temporal nature and change patterns in data, and significantly improve the accuracy and stability of rule mining. Experimental results show that our method has shown good performance on data set, especially in dynamic data environments, and can timely discover new rules and update existing rules.

Although our method has achieved good results in rule mining, there is still room for further optimization. Future research can focus on improving the computational efficiency of the algorithm, especially when dealing with large-scale data sets. How to improve the computational efficiency of the model while ensuring the accuracy of rule mining is a question worth exploring. In addition, as the complexity of data increases, our algorithm can be further combined with other deep learning models, such as graph neural networks (GNNs) or reinforcement learning algorithms, to enhance the performance of the model in high-dimensional and complex data.

Looking ahead, dynamic rule mining algorithms based on improved Transformers have broad application prospects. Especially in the fields of financial risk control, medical diagnosis, and intelligent recommendation, they can discover potential rules and trends through deep mining of dynamic data, thereby providing support for decision-making. With the continuous advancement of technology, we believe that such algorithms will be able to handle more complex and diverse data and provide more accurate solutions for intelligent decision-making in practical applications.